%% file: main.tex
\documentclass{Interspeech2024}

\interspeechcameraready

\title{Multimodal Belief Prediction}

\name[affiliation={\ast1,3}]{John}{Murzaku}
\name[affiliation={\ast1,3}]{Adil}{Soubki}
\name[affiliation={2,3}]{Owen}{Rambow}

\address{
  Stony Brook University, USA \\
  $^1$Department of Computer Science \\
  $^2$Department of Linguistics\\
  $^3$Institute for Advanced Computational Science 
}
\email{\{jmurzaku,asoubki\}@cs.stonybrook.edu, owen.rambow@stonybrook.edu}


\usepackage{xspace}

\newcommand{\tbf}[1]{\textbf{#1}}
\newcommand{\tss}[1]{\textsuperscript{#1}}

\makeatletter
\let\oldfootnote\footnote
\renewcommand\footnote[1]{%
  \begingroup
  \let\thefootnote\relax
  \oldfootnote{#1}%
  \addtocounter{footnote}{-1}%
  \endgroup
}
\makeatother

\newcommand{\mytilde}{\raise.17ex\hbox{$\scriptstyle\mathtt{\sim}$}}

\usepackage{colortbl}

\usepackage{todonotes}

\begin{document}

\maketitle

\begin{abstract}
Recognizing a speaker's level of commitment to a belief is a difficult task; humans do not only interpret the meaning of the words in context, but also understand cues from intonation and other aspects of the audio signal. Many papers and corpora in the NLP community have approached the belief prediction task using text-only approaches. We are the first to frame and present results on the multimodal belief prediction task. We use the CB-Prosody corpus (CBP), containing aligned text and audio with speaker belief annotations. We first report baselines and significant features using acoustic-prosodic features and traditional machine learning methods. We then present text and audio baselines for the CBP corpus fine-tuning on BERT and Whisper respectively. Finally, we present our multimodal architecture which fine-tunes on BERT and Whisper and uses multiple fusion methods, improving on both modalities alone. 
\end{abstract}

\section{Introduction}
A critical aspect of both written and spoken conversation is identifying the extent to which discourse participants 
commit themselves to propositions. However, as is frequently observed in sociolinguistics \cite{brown1987politeness}, humans rarely assert commitments exactly and employ all manner of tools to convey this tactfully. This makes the task of extracting and predicting how committed speakers are to propositions (i.e., their beliefs) more challenging than it might first seem, especially if we are given only written text. Previous work has shown that acoustic-prosodic speech features can distinguish deception (no commitment to a belief) and truthfulness (commitment to a belief), or trust and mistrust \cite{levitan18_interspeech,levitan2022believe,chen2020acoustic}, which emphasizes the need to jointly model text and audio features.
\footnote{$^\ast$Denotes equal contribution to this paper.}

Over the last 15 years, the belief prediction task (also called event factuality prediction) has attracted significant interest from the NLP community, resulting in several corpora \cite{sauri2009factbank,prabhakaran-etal-2015-new,minard-etal-2016-meantime,white-etal-2018-lexicosyntactic,rudinger-etal-2018-neural-models,de2019commitmentbank,markowska-etal-2023-finding}. However, most of this work has only focused on predicting belief from the text modality. A notable exception is the CommitmentBank (CB) corpus \cite{de2019commitmentbank} which annotates both text and audio corpora, with 350 utterances coming from Switchboard
\cite{stolcke-etal-2000-dialogue}. However, the annotators only read the dialog transcription and audio was discarded. Expanding on CB, the authors present the CB-Prosody (CBP) corpus \cite{mahler2020prosody}, which annotates speaker commitment/belief on the same 350 Switchboard examples, but instead annotators only hear the audio. To our knowledge, we are the first to present a multimodal belief prediction task and find that incorporating audio signals results in a 12.7\% relative reduction in mean absolute error (MAE) and 6.4\% relative increase in Pearson correlation when compared to text-only approaches.\oldfootnote{\scriptsize\url{https://github.com/cogstates/multimodal-belief}}

Our main contributions are summarized as follows:
\begin{enumerate}
    \item We are the first to present and report results on the multimodal belief prediction task and do so using the CBP corpus. Furthermore, we contribute missing information to the CBP corpus (audio start and end times) and will release all data. 
    \item We perform an acoustic-prosodic analysis of CBP using openSMILE features \cite{opensmile}, and find 25 significant features. We train the acoustic-prosodic features using a XGBoost-RF model and are the first to report audio-only results with these features. 
    \item We present audio-only and text-only baselines with pre-trained models, specifically fine-tuning BERT \cite{devlin-etal-2019-bert} and Whisper \cite{radford2023robust}. We find that Whisper significantly outperforms our acoustic-prosodic XGBoost-RF baseline. 
    \item We present a multimodal belief architecture, jointly fine-tuning BERT and Whisper. We compare our multimodal results to our text-only and audio-only baselines, and ultimately find our multimodal architecture provides a significant improvement in belief detection. Furthermore, we investigate early and late fusion methods, finding that late fusion of features performs better than early fusion. 
\end{enumerate}
The paper is organized as follows. Section~\ref{sec:data} describes our data, specifically the CBP corpus and how it frames the belief prediction task. In Section~\ref{sec:approaches}, we describe our acoustic-prosodic and neural approaches to the task. Results for all modalities and experimental settings are reported in Section~\ref{sec:experiments}. We conclude in Section~\ref{sec:conclusion} and present ideas for future work in multimodal belief detection.\looseness=-1


\section{CB-Prosody Corpus} \label{sec:data}
We use the CB-Prosody corpus (CBP), which annotates on top of the manually transcribed English Switchboard corpus utterances.
Annotators were given speaker utterances, specifically the content of the complement clause, and asked to evaluate the level of certainty the speaker appears to have regarding the truth of the proposition. 
Specifically, they were asked to rate the level of commitment on a 7-point Likert scale where explicit labels were displayed for three classes: the speaker is certain that it is true, the speaker is not certain whether it is true or false, and the speaker is certain that it is false. The intermediate values contained partial commitments (i.e. the speaker is weakly or strongly possibly certain/not certain).
The average among at least eight annotators' judgements
is then reported as the final annotation value.

We manually extract the clip start and end times from Switchboard, convert the files to wav format, and then perform upsampling to 16kHz on the audio clips. We discard 12 examples that were either misaligned with the gold Switchboard files or had significant interruptions (not a continuous utterance from the speaker) and end up with 338 total examples. 

Summary statistics for the audio segments and word counts in each example are shown in Table~\ref{tab:dataset_stats}.
\begin{figure}

    \centering
    \includegraphics[width=\linewidth]{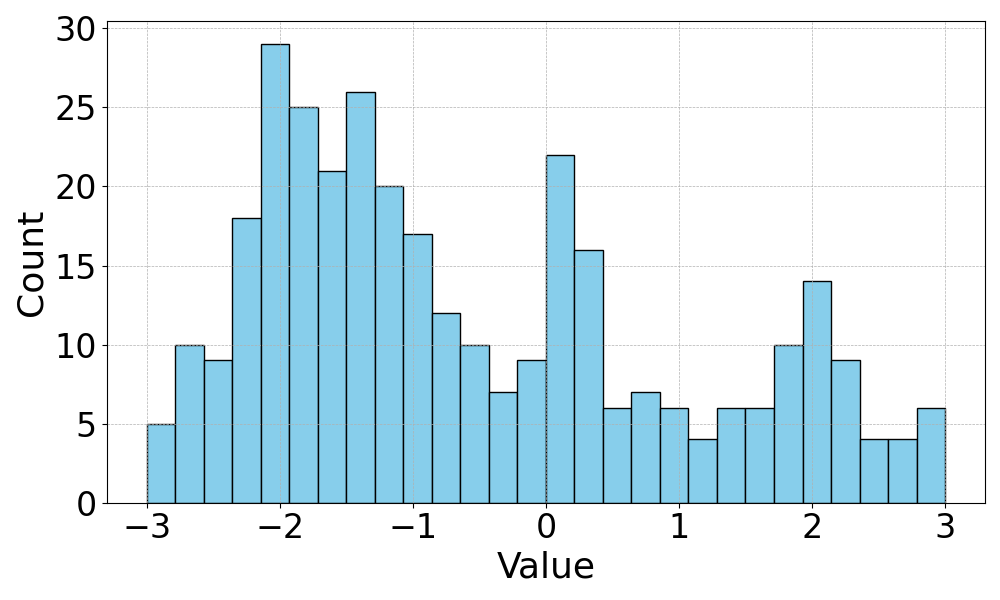}
        \caption{A histogram of the average commitment values among annotators for all 338 utterances in CB Prosody.}
    \label{tab:label_dist}
    
\end{figure}
All annotations are provided on a continuous scale. To report dataset statistics, we discretize the continuous annotations into 28 bins.
Figure~\ref{tab:label_dist} shows our bins and the corresponding count for each bin. There is a clear imbalance in the label distribution with the majority of labels falling in between the bin values $[-2, -1]$. There are also two noticeable peaks around bin values 0 and 2. The fewest label counts are near bin values -3, 1, and 3. 

\begin{table}[h]
\caption{CBP summary statistics}
\centering
\begin{tabular}{|c|c|c|c|c|c|}
\hline
 & Min & Max & Mean & Median & Stdv \\ \hline
Duration (s) & 1.01 & 16.44 & 4.61 & 4.04 & 2.73 \\ \hline
Words & 5.00 & 47.00 & 16.49 & 15.00 & 7.48 \\ \hline
\end{tabular}
\label{tab:dataset_stats}
\end{table}

To illustrate the difference between the text-only speaker commitment annotations on CB and the audio speaker commitment annotations on CBP, we consider the following example from the corpus: \emph{He didn't think it looked predawn to him.} In this example, annotators are asked to annotate how certain the speaker is that \emph{it looked predawn to him}. We note that the speaker is not the referent of {\em he} and {\em him} -- this corpus does not annotate the cognitive state of people mentioned in text, only of the speaker. Here, the text-only CB corpus annotates this as 0.55, suggesting a slight inclination towards non-commitment (i.e., 0) regarding the statement about it looking predawn. To an annotator who only sees the text, this leaning towards no commitment is due to the lexical content and syntactic structure of the sentence, where the negation (\textit{He didn't think}) could introduce uncertainty about the speaker's commitment. 

On the other hand, the CBP corpus annotates this as -2.16, indicating a strong commitment to the falseness of the statement (i.e., committing to it \textit{not looking predawn}). In this example's audio, there is a heavy emphasis on \emph{he didn't}, suggesting that the speaker trusts the “he”, and this he believes that it doesn't look predawn to him, meaning a strong leaning towards committing to the falseness of the statement. 



\section{Our Approaches} \label{sec:approaches}

We frame belief prediction as a regression task, following previous literature \cite{rudinger-etal-2018-neural-models,murzaku-etal-2022-examining,jiang-de-marneffe-2021-thinks,stanovsky-etal-2017-integrating}.
Given an utterance and a sequence of corresponding features $S = [f_1, f_2, . . . , f_n]$, we wish to produce a value $\hat{y} \in [-3, 3]$ that is as close to the annotation value $y$ as possible.

\begin{figure*}[t]
    \centering
    \vspace{-0.5em}
    \includegraphics[
        width=0.78\textwidth,
        height=0.22\paperheight
    ]{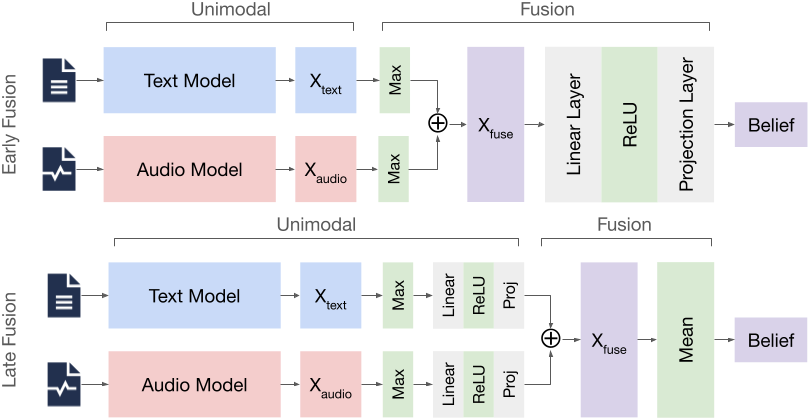}
    \caption{Block diagram describing the early (above) and late (below) fusion models.}
    \label{fig:architecture}
\end{figure*}


\begin{table}[t]
\centering
\caption{Statistically significant top 11 openSMILE IS09 features}
\begin{tabular}{|c|c|c|}
\hline
Rank & Feature & Avg $r$ \\ \hline
1-5 & MFCC & 0.158 \\ 
6-7 & F0 & 0.124 \\ 
8-10 & MFCC & 0.120 \\ 
11 & voiceProb & 0.112 \\ \hline
\end{tabular}
\label{tab:top_features}
\end{table}

\subsection{Audio Feature Based}
For our feature based approach, we extract features from our audio segments using openSMILE \cite{opensmile}. We hypothesize that there is an overlap in significant features between our task and deception detection. Hence, we specifically use the 384 feature Interspeech 2009 (IS09) emotion challenge set \cite{schuller2009interspeech}, which previous work on deception detection found to be useful \cite{levitan18_interspeech,amiriparian16_interspeech}. 

We also perform a simple feature selection approach and aim to answer the following question: \textbf{Are there OpenSmile features (from the IS09 set) that are correlated with the continuous CBP annotation values?} To answer this, we calculate the Pearson $\rho$ among our features and the belief values and choose the statistically significant features (any correlation with p-value $<$ 0.05).
We find 25 statistically significant features and provide a summary of our top 11 features (measured by p-value) in Table 2. The most represented features are MFCC measures/coefficients, showing in 8 of our top 11 features. We also find F0 and voiceProb measures/coefficients as significant features. Interestingly, our significant features are directly in line with
significant features found in deception detection such as MFCCs and voiceProb \cite{Mendels2017HybridAD}, which points to the similarities between our tasks.


\subsection{Fine-tuning}
The models we fine-tuned for each modality are as follows:

\noindent \textbf{Text} --
We
use the pre-trained BERT model \cite{devlin-etal-2019-bert} for our fine-tuning experiments, specifically \textit{bert-base-uncased} which contains 110M parameters. We also performed fine-tuning experiments with
RoBERTa \cite{liu2019roberta} and got near identical or worse results which is in line with previous work on CB \cite{jiang-de-marneffe-2021-thinks}, so we therefore use BERT instead. We follow a standard fine-tuning approach and 
add a regression head to which we feed the final BERT representations to predict a continuous value. 

\noindent \textbf{Audio} -- We use the pre-trained Whisper model \cite{radford2023robust} for our fine-tuning experiments, specifically \textit{whisper-base}, which contains 72.6M parameters. We note that we also tried experiments with Hubert \cite{hubert}
, but achieved better performance with Whisper. We try both mean and max pooling of our audio features after fine-tuning and before our regression head, finding that max pooling slightly outperforms mean pooling. We therefore perform all experiments with max pooling.
Similarly to our text only experiments, we perform a standard fine-tuning approach, only adding a regression head to predict a continuous value, to which we feed the final Whisper representations to.

\noindent \textbf{Multimodal} -- Our multimodal architecture fine-tunes on both BERT and Whisper. A natural research question that arises for our task is which multimodal feature fusion method performs best: early feature fusion or late feature fusion. In the multimodal emotion recognition task, which uses similar methods of jointly fine-tuning of text and audio pre-trained models, it has been shown that late feature fusion works better than early feature fusion \cite{zhao22k_interspeech}. A similar result was obtained for multimodal persuasion prediction \cite{nojavanasghari2016deep}. We describe our feature fusion architectures in the next subsections.

All of our fine-tuning experiments are run on a single Tesla V100 or H100 GPU. Each fold took on average about 15 minutes to fine-tune, totaling about 75 minutes per experiment configuration. 

\subsection{Multimodal Early Fusion} Our early fusion model
is shown at the top of Figure~\ref{fig:architecture}. We first jointly fine-tune BERT and Whisper on our text and audio inputs, respectively. 
We then pass the final hidden representations from each modality through a max pooling layer and concatenate them. 
Finally, we pass the concatenated pooled representations to our regression head consisting of a linear layer, ReLU, and output a final continuous value. When incorporating openSMILE features, they are concatenated after pooling.

\subsection{Multimodal Late Fusion}
Our late fusion model is shown at the bottom of Figure~\ref{fig:architecture}. We first fine-tune BERT and Whisper on our text and audio inputs, respectively. We do not concatenate them as we do in our early fusion model. Rather, we max pool both representations individually and then pass them through their individual regression heads. We then finally concatenate the two representations
and do mean averaging to get our final continuous value. When using openSMILE features, they are passed through their own regression head.

\section{Experiments and Discussion} \label{sec:experiments}
\begin{table}[t]
\caption{XGB-RF results with all IS09 features and correlated IS09 features. }
\centering
\begin{tabular}{|c|c|c|}
\hline
 Features & MAE $\downarrow$ & Pearson $\uparrow$ \\ \hline
All IS09 Feats & 1.25\tiny$\pm$\tiny0.09  & 0.18\tiny$\pm$\tiny0.10 \\ \hline
Correlated Feats & 1.22\tiny$\pm$\tiny0.08  &  0.24\tiny$\pm$\tiny0.04 \\ \hline
\end{tabular}
\label{tab:feats_xgbrg}
\end{table}

\subsection{Data}
We
use all 338 examples of the CBP corpus described in Section~\ref{sec:data}. To address the label imbalance shown in Figure 1, we perform a stratified five-fold cross validation using a fixed seed for all experiments.
 We do not hold out an established validation set because of the small size of the corpus, and therefore do not perform any hyperparameter tuning.

\subsection{Training and Evaluation Methods}
For our tasks, we either directly perform regression or add a regression head to our pre-trained models. Our approaches are as follows:

\noindent \textbf{XGB-RF} For our audio only feature based approach with openSMILE, we train an XGBoost-Random Forest (XGB-RF) regression model. We perform z-score normalization on all of our features. We do not perform hyperparameter tuning and use the standard XGB-RF hyperparameters of scikit-learn \cite{sklearn_api}.

\noindent \textbf{Fine-tuning} We fine-tune all models for a fixed 10 epochs
and report the Pearson correlation and mean absolute error at the last epoch. All experiments use a learning rate of 2e-5
and a batch size of 1. We do not perform any hyperparameter tuning or hyperparameter searches. We use the mean squared error (MSE) loss function. We note that we tried the Huber loss function which was used in previous work on text-only belief prediction \cite{jiang-de-marneffe-2021-thinks,murzaku-etal-2022-examining}, but ultimately found that this performed worse compared to MSE. For our BERT experiments, we set the maximum sequence length to 256. For our Whisper experiments, we pad all audio clips to 30 seconds and perform normalization.

\noindent \textbf{Evaluation} We
evaluate on Pearson correlation ($\rho$) and mean absolute error (MAE). Previous work on text-only belief prediction found that Pearson $\rho$ is useful in biased test sets since it assesses how well a model captures variability in the gold data and MAE captures absolute fit \cite{stanovsky-etal-2017-integrating}. We report the average Pearson $\rho$ and MAE across all five folds and their corresponding standard deviations. We perform significance testing on both metrics using paired bootstrap resampling, which has shown to be notably effective and robust for small test sets \cite{koehn2004statistical}.

\input{figures/results.tex}

\subsection{Unimodal and Multimodal Models}
We first train our XGB-RF regressor for our acoustic-prosodic feature baseline on all the IS09 features and then only the significant IS09 features described in Section 3. Results are shown in Table 3. We find that using correlated features results in a 2.4\% decrease in MAE and a 33.3\% increase in Pearson $\rho$.

The results for all model configurations are shown in Table \ref{tab:results}. The best unimodal model is text-only BERT which achieves a mean MAE of 0.71 and Pearson $\rho$ of 0.78. While the remaining four audio-only unimodal models tested consistently perform worse than the text-only model, the multimodal fusion models decidedly outperform unimodal approaches. Our best result comes from the late fusion of BERT and Whisper and significantly ($p<0.05$) improves over unimodal BERT achieving a mean MAE of 0.62 and Pearson $\rho$ of 0.83. These are 12.7\% and 6.4\% relative improvements in MAE and Pearson, respectively. In fact, our three best results come from multimodal fusion approaches and all result in significant ($p<0.05$) improvements in MAE demonstrating the value of incorporating audio signals for belief prediction.

\subsection{Impact of Fusion Strategy}
Theoretical work comparing early and late fusion methods predicts that, with enough training data, early fusion will perform best but, in the absence of sufficient data, late fusion will do better \cite{pereira2023comparing}. Comparing our models by fusion strategy we see that, as theory predicts given our small dataset, the late fusion models tend to perform better regardless of the features being incorporated. While we do not find significant differences (MAE $p=0.11$, Pearson $p=0.17$) between early and late fusion of BERT and Whisper, we do find significant differences when we compare all early fusion results with all late fusion results.
This pattern has also been observed for related tasks including persuasiveness prediction ($n \approx 205$) \cite{nojavanasghari2016deep} and emotion recognition ($n \approx 4,400$) \cite{zhao22k_interspeech}. Our results support existing literature both theoretically and empirically.

\subsection{Comparison of Whisper and openSMILE}
It is not entirely clear that the pre-training objective used for Whisper will produce representations which capture information similar to the acoustic-prosodic features from openSMILE. It is possible that Whisper and openSMILE provide some orthogonal information. When comparing late fusion of BERT and Whisper with the late fusion of BERT and openSMILE we find that Whisper significantly ($p < 0.05$) outperforms openSMILE as a source of audio features by MAE (though not by Pearson). 
When comparing to the late fusion of all three feature sources, we do not find the increase compared to our BERT text-only model
to be significant for Pearson $\rho$. However, the MAE of fusing all three features
was lower and significant. This suggests that Whisper is the best source of audio features, though openSMILE performs competitively and may be preferable in applications with limited computational resources, or in situations in which we need to use interpretable features.

Incorporating openSMILE using early fusion performed much worse than by using late fusion. While this was expected, as discussed above, it was more substantial than observed with Whisper. We suspect it is possible to reduce this gap with additional changes to the architecture or normalization techniques but leave this to future work as our primary objective is to demonstrate the importance of audio features for this task.

\subsection{Comparison to Other Work}
While there are no comparable results on CBP, text-only belief prediction is a well established task in NLP. For the entire CB corpus, containing 556 utterances,
state-of-the-art results achieve an MAE of 0.56. Given the disparity in training examples, our best performing model does well with an MAE of 0.62 (and 0.71 text-only).
Performance for the text-only task varies among other belief corpora with state-of-the-art results ranging from 0.33 MAE on FactBank to 0.73 on UDS-IH2 \cite{rudinger-etal-2018-neural-models, murzaku-etal-2022-examining}. 

Though none of these make use of audio signals, our results are roughly in line with them which suggests that our approach achieves strong performance even in this broader framing.

\section{Conclusion and Future Work} \label{sec:conclusion}
In this paper, we are the first to present audio-only and multimodal results on the belief detection task. We first analyze acoustic-prosodic features extracted from openSMILE, and provide an audio only baseline. We find similarities and overlaps with the deception detection task. Next, we provide unimodal and multimodal baselines fine-tuning pre-trained speech and text models. For our audio only baseline, we find that fine-tuning Whisper outperforms our acoustic-prosodic feature-based approach with openSMILE. We explore early and late fusion techniques, and find both early fusion and late fusion help compared to text-only. We achieve a new state-of-the-art result by fine-tuning BERT and Whisper, and using late fusion. Our state-of-the-art system achieves a 12.7\% decrease in MAE and a 6.4\% increase in Pearson $\rho$ compared to our text-only baseline.  

Our results provide a baseline for the multimodal belief prediction task and show that an audio signal does help text-only models. With these insights, in future work we intend to apply multimodal belief models on other corpora such as the Common Ground corpus \cite{markowska-etal-2023-finding} which annotates beliefs and common grounds  on the CALLHOME corpus \cite{CallhomeCorpus}. 
We intend to do multi-task learning experiments on belief and deception and see if they can help each other. 

\bibliographystyle{IEEEtran}
\bibliography{mybib}

\section{Acknowledgements}
This material is based upon work supported in part by the National Science Foundation (NSF) under No. 2125295 (NRT-HDR: Detecting and Addressing Bias in Data, Humans, and Institutions) and by funding from the Defense Advanced Research Projects Agency (DARPA) under the INCAS (HR01121C0186, No. HR001120C0037, and PR No. HR0011154158) and the CCU (No. HR001120C0037, PR No. HR0011154158, No. HR001122C0034) programs. Any opinions, findings and conclusions or recommendations expressed in this material are those of the author(s) and do not necessarily reflect the views of the NSF or DARPA. 

We thank the Institute for AI-Driven Discovery and Innovation at Stony Brook for access to the computing resources needed for this work made possible by NSF grant No. 1919752 (Major Research Infrastructure program).
\end{document}

%% file: figures/results.tex
\begin{table}[tb]
\setlength\tabcolsep{5.78pt}

\caption{Mean MAE and Pearson $\rho$ results for the 5-fold CV experiments comparing modality and fusion strategies. Significant ($p<0.05$) improvements  over the text-only BERT model are indicated by $\dagger$.}
\centering
\begin{tabular}{lll|cc}
    \hline
    \tbf{Text} & \tbf{Audio} & \tbf{Fusion} &
    \tbf{MAE $\downarrow$} & \tbf{Pearson$\uparrow$} \\
    \hline
    BERT & -         & -      
                     & {\cellcolor[HTML]{BFBFBF}} \color[HTML]{000000} 0.71 \tiny$\pm$\tiny 0.08 
                     & {\cellcolor[HTML]{BFBFBF}} \color[HTML]{000000} 0.78 \tiny$\pm$\tiny 0.03 \\ 
    -    & Whisper   & -      
                     & {\cellcolor[HTML]{F0F0F0}} \color[HTML]{000000} 1.02 \tiny$\pm$\tiny 0.14 
                     & {\cellcolor[HTML]{F0F0F0}} \color[HTML]{000000} 0.53 \tiny$\pm$\tiny 0.13 \\
    -    & OpenSMILE & -      
                     & {\cellcolor[HTML]{F0F0F0}} \color[HTML]{000000} 1.22 \tiny$\pm$\tiny 0.08 
                     & {\cellcolor[HTML]{F0F0F0}} \color[HTML]{000000} 0.24 \tiny$\pm$\tiny 0.04 \\
    -    & Whisper + OS & Early 
                        & {\cellcolor[HTML]{F0F0F0}} \color[HTML]{000000} 1.27 \tiny$\pm$\tiny 0.07 
                        & {\cellcolor[HTML]{F0F0F0}} \color[HTML]{000000} 0.14 \tiny$\pm$\tiny 0.10 \\
    -    & Whisper + OS & Late  
                        & {\cellcolor[HTML]{F0F0F0}} \color[HTML]{000000} 1.10 \tiny$\pm$\tiny 0.15 
                        & {\cellcolor[HTML]{F0F0F0}} \color[HTML]{000000} 0.41 \tiny$\pm$\tiny 0.19 \\
    \hline
    BERT & Whisper   & Early  
                     & {\cellcolor[HTML]{808080}} \color[HTML]{F1F1F1} 0.66\tss{$\dagger$}\hspace{-0.5em} \tiny$\pm$\tiny 0.09 
                     & {\cellcolor[HTML]{808080}} \color[HTML]{F1F1F1} 0.81\tss{$\dagger$}\hspace{-0.5em} \tiny$\pm$\tiny 0.04 \\
    BERT & Whisper   & Late   & {\cellcolor[HTML]{404040}} \color[HTML]{F1F1F1} 0.62\tss{$\dagger$}\hspace{-0.5em} \tiny$\pm$\tiny 0.08
                              & {\cellcolor[HTML]{404040}} \color[HTML]{F1F1F1} 0.83\tss{$\dagger$}\hspace{-0.5em} \tiny$\pm$\tiny 0.04 \\
    BERT & OpenSMILE & Early  
                     & {\cellcolor[HTML]{F0F0F0}} \color[HTML]{000000} 1.19 \tiny$\pm$\tiny 0.06 
                     & {\cellcolor[HTML]{F0F0F0}} \color[HTML]{000000} 0.37 \tiny$\pm$\tiny 0.06 \\
    BERT & OpenSMILE & Late   
                     & {\cellcolor[HTML]{A0A0A0}} \color[HTML]{F1F1F1} 0.68 \tiny$\pm$\tiny 0.10 
                     & {\cellcolor[HTML]{A0A0A0}} \color[HTML]{F1F1F1} 0.80 \tiny$\pm$\tiny 0.04 \\
    BERT & Whisper + OS & Early  
                        & {\cellcolor[HTML]{F0F0F0}} \color[HTML]{000000} 0.92 \tiny$\pm$\tiny 0.05 
                        & {\cellcolor[HTML]{F0F0F0}} \color[HTML]{000000} 0.62 \tiny$\pm$\tiny 0.10 \\
    BERT & Whisper + OS & Late   & {\cellcolor[HTML]{808080}} \color[HTML]{F1F1F1} 0.66\tss{$\dagger$}\hspace{-0.5em} \tiny$\pm$\tiny 0.08 
                                 & {\cellcolor[HTML]{A0A0A0}} \color[HTML]{F1F1F1} 0.79 \tiny$\pm$\tiny 0.03 \\
    \hline
\end{tabular}
\label{tab:results}
\end{table}